\documentclass[letterpaper, 10 pt, conference]{ieeeconf}  

\IEEEoverridecommandlockouts

\usepackage{amsmath} 
\usepackage{hyperref}
\usepackage{amssymb} 
\usepackage{dsfont}
\usepackage{graphicx}
\usepackage{subcaption}
\usepackage{psfrag,graphicx,epsfig}
\usepackage{epstopdf}
\usepackage{xspace}
\usepackage{float}
\usepackage{placeins}
\usepackage{multirow}
\usepackage{pgf,tikz}
\usepackage{nowidow}
\usepackage{lineno}
\usepackage{xcolor}
\usepackage{siunitx}
\usepackage{color}
\usepackage{makecell}
\usepackage{lineno}
\usepackage{overpic}
 \usepackage{booktabs}

\newcommand{\unic}{\textsc{UNIC}}

\title{\LARGE \bf

UNIC: Learning Unified Multimodal Extrinsic Contact Estimation
}

\author{Zhengtong Xu and Yuki Shirai  
\thanks{Zhengtong Xu is with the Edwardson School of Industrial Engineering, Purdue University, West Lafayette, USA  {\tt\small xu1703@purdue.edu}}
\thanks{Yuki Shirai is with Mitsubishi Electric Research Laboratories, Cambridge, MA, USA {\tt\small shirai@merl.com}}
}

\begin{document}

\maketitle
\thispagestyle{empty}
\pagestyle{empty}


\begin{abstract}

Contact-rich manipulation requires reliable estimation of extrinsic contacts—the interactions between a grasped object and its environment—which provide essential contextual information for planning, control, and policy learning. However, existing approaches often rely on restrictive assumptions, such as predefined contact types, fixed grasp configurations, or camera calibration, that hinder generalization to novel objects and deployment in unstructured environments. In this paper, we present \unic{}, a unified multimodal framework for extrinsic contact estimation that operates without any prior knowledge or camera calibration. \unic{} directly encodes visual observations in the camera frame and integrates them with proprioceptive and tactile modalities in a fully data-driven manner. It introduces a unified contact representation based on scene affordance maps that captures diverse contact formations and employs a multimodal fusion mechanism with random masking, enabling robust multimodal representation learning.

Extensive experiments demonstrate that \unic{} performs reliably. It achieves a 9.6 mm average Chamfer distance error on unseen contact locations, performs well on unseen objects, remains robust under missing modalities, and adapts to dynamic camera viewpoints. These results establish extrinsic contact estimation as a practical and versatile capability for contact-rich manipulation. The overview and hardware experiment videos are \href{https://youtu.be/xpMitkxN6Ls?si=7Vgj-aZ_P1wtnWZN}{here}.
\end{abstract}

\section{Introduction}\label{sec:intro}
Extrinsic contact \cite{ma2021extrinsic} refers to contact events involving external objects and the environment, beyond the robot’s own body. Estimating extrinsic contact is particularly challenging \cite{rodriguez2021unstable}, as it requires reasoning about interactions that are often indirect, involve multiple objects, and occur under uncertain geometry or dynamics. Nevertheless, extrinsic contact estimation is essential for enhancing robotic dexterity, as it enables robots to perceive and reason about diverse interactions with the environment. From tool use \cite{kim2024texterity,shirai2023tactile} and non-prehensile manipulation \cite{billard2019trends} to precise environment interactions \cite{higuera2023perceiving}, reliable extrinsic contact estimation supports robust performance while ensuring safety and stability in diverse, unstructured settings.

Recent research has investigated tactile-based and multimodal approaches for extrinsic contact estimation. However, these methods often depend on strong priors or restrictive assumptions, such as predefined contact types \cite{ma2021extrinsic,ota2024tactile,oller2024tactile}, initial contact conditions \cite{kim2023simultaneous}, fixed grasps without slip \cite{higuera2023neural}, object geometries \cite{lee2025vitascope}, camera calibration \cite{lee2025vitascope,oller2024tactile}, or fixed camera placement \cite{yi2024visual}. These constraints limit their practicality in real-world scenarios, hinder the development of a unified framework, and restrict deployment in diverse, unstructured environments. Moreover, these limitations diminish the feasibility of leveraging extrinsic contact estimation as a general-purpose capability for downstream applications, such as integrating contact estimation with policy learning.

\begin{figure}[t]
\centering
\begin{overpic}[trim=0 0 0 0,clip, width=0.5\textwidth]{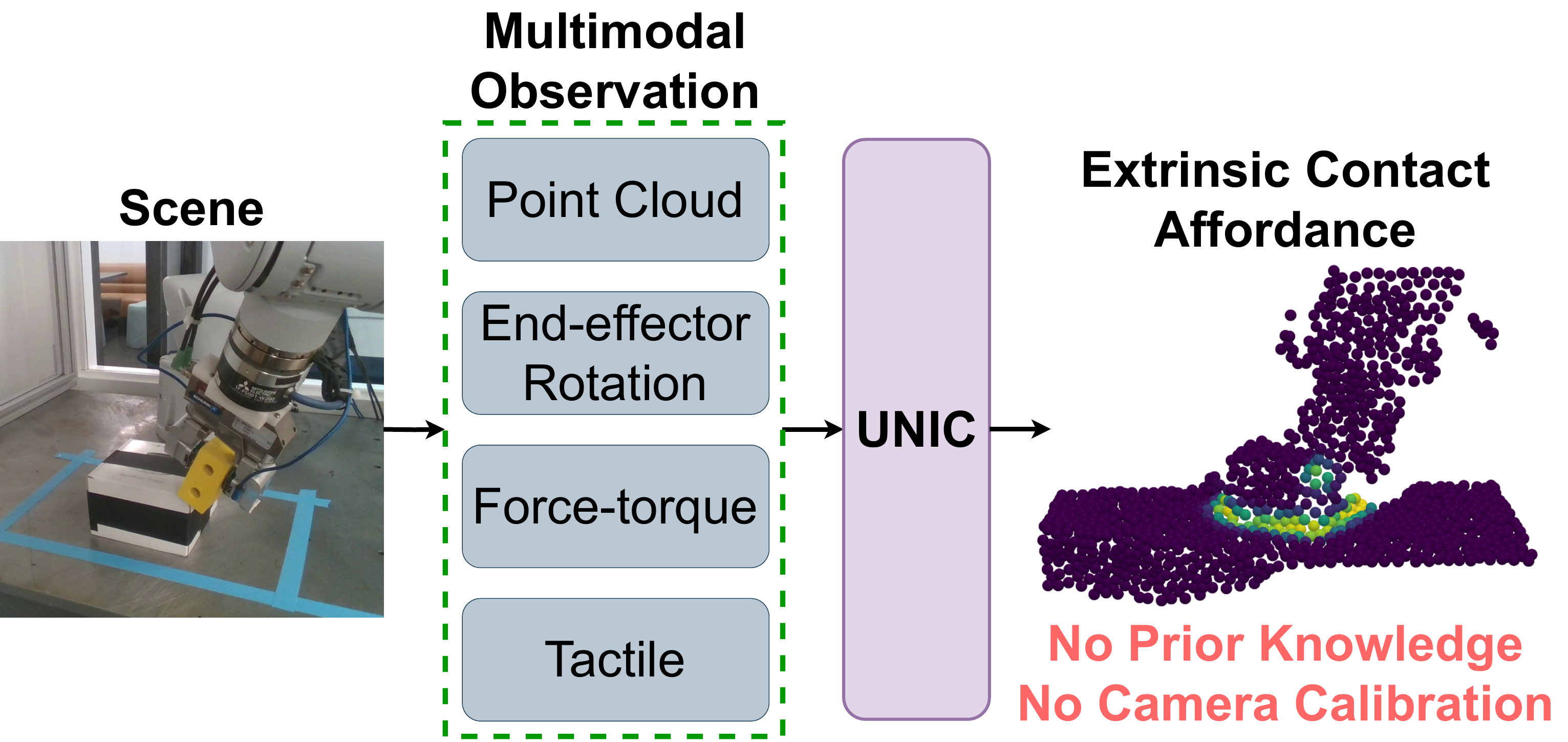}
\end{overpic}
\caption{\unic{} leverages multimodal inputs to estimate a unified contact affordance map that captures diverse forms of extrinsic contact, including complex interaction chains such as gripper–object–object–environment. Throughout this process, \unic{} does not rely on prior knowledge or camera calibration. At deployment, it remains effective even under missing modalities, adapts to dynamic camera viewpoints, and generalizes well to unseen objects. The overview and hardware experiment videos are \href{https://youtu.be/xpMitkxN6Ls?si=7Vgj-aZ_P1wtnWZN}{here}.}
\label{fig:first_page}
\end{figure}

In this paper, we present \unic{}, a unified framework for extrinsic contact estimation, as shown in Fig.~\ref{fig:first_page}. The main contributions of this work are as follows:

1. \textbf{Prior-free framework:} We present a unified framework for extrinsic contact estimation that requires neither prior knowledge nor camera calibration. Visual observations are encoded directly in the camera frame and fused with proprioceptive and tactile inputs in a fully data-driven manner. This design supports flexible deployment under varying camera viewpoints and generalizes effectively to diverse scenarios, including previously unseen objects.

2. \textbf{Unified contact representation:} We introduce a unified contact representation based on scene affordance maps, which captures transitions from no-contact to contact states and accommodates diverse contact types, including point, line, and patch. Moreover, this representation explicitly models complex contact chains, extending beyond grasped object–environment interactions to encompass contacts among multiple objects in the scene.

3. \textbf{Unified multimodal fusion:} We propose a unified multimodal fusion mechanism that randomly masks feature tokens during training, encouraging the model to learn robust cross-modal representations. Consequently, \unic{} can deliver reliable estimation even when one or more modalities are missing at deployment, thereby greatly improving the flexibility and practicality of real-world deployment.

\section{Related Work}

\begin{figure*}[h]
\centering
\begin{overpic}[trim=0 0 0 0,clip, width=1\textwidth]{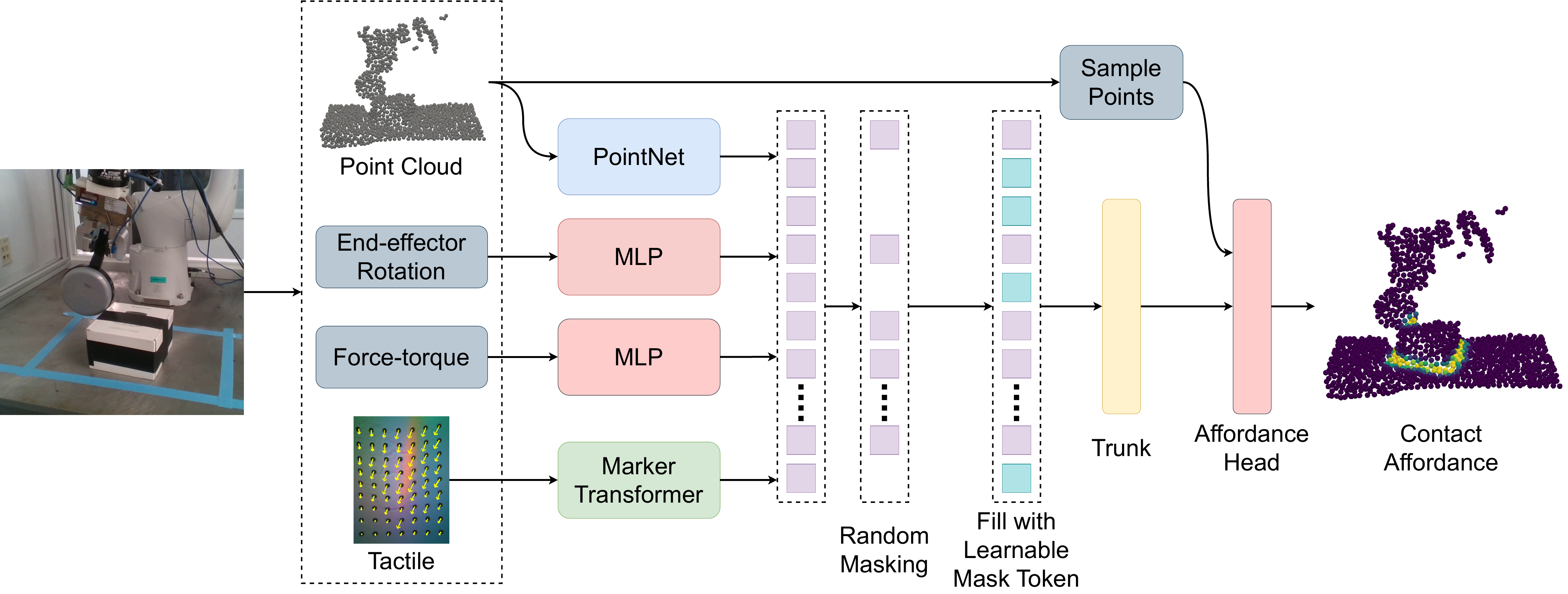}
\end{overpic}
\caption{Pipeline of \unic{}. \unic{} integrates four sensing modalities as inputs—point clouds, end-effector rotation, force–torque, and tactile marker displacements—and outputs an extrinsic contact affordance map. We adopt a masked multimodal fusion strategy to ensure a robust multimodal representation learning. In addition, \unic{} employs a sampling strategy designed to enhance computational efficiency. }
\label{fig:pipeline}
\end{figure*}

In this section, we review prior work on extrinsic contact estimation, which can be broadly categorized into model-based and model-free methods.

Model-based methods rely heavily on underlying physical models \cite{ma2021extrinsic, shirai2023tactile, taylor2023object, oller2024tactile, kim2024texterity, bianchini2025vysics}. These approaches collect data during manipulation and apply optimization techniques subject to physical constraints to infer unobservable contacts. For example, \cite{shirai2023tactile} employs nonlinear optimization with tactile sensing to estimate slip during tool manipulation, while \cite{kim2024texterity} leverages factor graphs with tactile input to estimate contacts.

Model-free methods \cite{higuera2023neural, yi2024visual, lee2025vitascope, ota2024tactile, aoyama2025few, fu2025unitac} directly map multimodal sensory inputs to contact states using neural networks. For instance, \cite{lee2025vitascope} trains a vision–tactile estimator in simulation and validates it in real settings, and  \cite{fu2025unitac} infers contact solely from joint torques and positions.

While these approaches have demonstrated promising results, they often depend on strong priors or restrictive assumptions that limit their practical deployment. Examples include predefined contact types \cite{ma2021extrinsic, ota2024tactile, oller2024tactile}, assumptions on initial contact conditions \cite{kim2023simultaneous}, fixed grasp configurations without slip \cite{higuera2023neural}, access to precise object geometries \cite{lee2025vitascope}, strict camera calibration \cite{lee2025vitascope, oller2024tactile}, or fixed viewpoints \cite{yi2024visual}.

In contrast, our work introduces a unified, prior-free framework for extrinsic contact estimation that is robust, deployable, and generalizable across diverse settings.

A related direction bypasses explicit contact estimation by learning end-to-end policies that map multimodal inputs directly to actions \cite{hansen2022visuotactile, qi2023general, xue2025reactive, liu2025factr, luu2025manifeel, xu2025unit}. While effective, these policies lack interpretability. In addition, they are task-specific, and require retraining for new tasks. In contrast, our task-agnostic estimator generalizes across conditions and provides a reusable capability that can interface with diverse manipulation policies.

\section{Method}
In this section, we present the details of \unic{}.

\subsection{Background: Extrinsic Contact}

Extrinsic contact requires reasoning about interactions such as gripper–object, object–object, and object–environment, where object geometry and pose are often unknown \cite{ma2021extrinsic,rodriguez2021unstable}. 

Previous work defined extrinsic contact mainly as gripper–object–environment interactions. Our formulation generalizes this to a unified representation that also captures complex cases such as multi-object manipulation (e.g., gripper–object–object–environment), where contacts may occur simultaneously across multiple objects and the environment.

\subsection{Overview of \unic{}}

The architecture of \unic{} is illustrated in Fig.~\ref{fig:pipeline}. The objective of \unic{} is to estimate extrinsic contact by \textbf{(i)} leveraging multimodal inputs, including point clouds, tactile signals, force–torque, and end-effector rotation, and \textbf{(ii)} avoiding reliance on additional information such as pre-constructed object geometries or camera calibration.

\begin{figure*}[h]
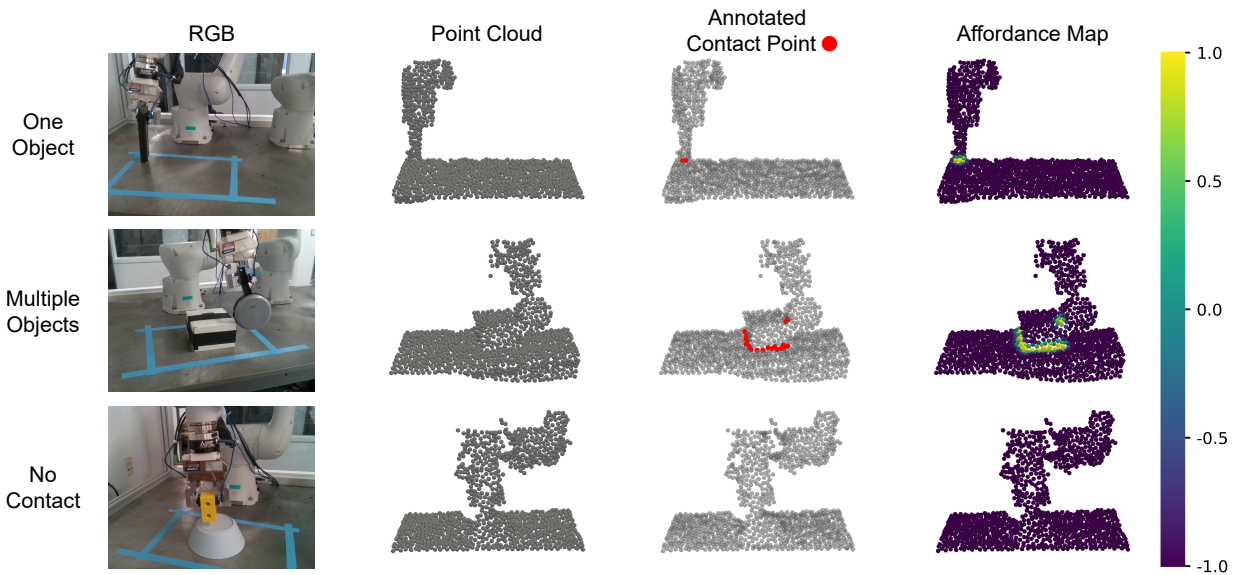

\centering
\begin{overpic}[trim=0 0 0 0,clip, width=0.93\textwidth]{figures/contact_representation.pdf}
\end{overpic}
\caption{Illustration of proposed prior-free contact affordance representation over three different cases. 
For each case, we show the RGB image and the corresponding point cloud. The human operator annotates the contact points. Based on these annotations, the proposed Gaussian kernel-based generation method is applied to produce the contact affordance map. Affordance values range from $-1$ to $1$, with higher values at contact locations and lower values in non-contact regions.}
\label{fig:contact_representation}
\end{figure*}

\subsection{Prior-free Contact Affordance Representation}

We introduce our prior-free contact affordance representation, a unified formulation capable of capturing diverse contact types, including complex chains such as gripper–object–object–environment, without requiring priors like camera calibration or pre-defined object geometries.

We represent extrinsic contact as an affordance map over the entire scene. As shown in Fig.~\ref{fig:contact_representation}, \unic{} directly uses the camera-frame point cloud as the reference for constructing this affordance map, without relying on ground-truth information such as the object’s geometries. Although this design inevitably sacrifices some precision and completeness due to the limited resolution and coverage of point clouds, it eliminates the need for constraints such as camera calibration and ensures that the learning process remains independent of additional priors from multimodal sensory inputs. This paradigm enables flexible deployment, allowing cameras to be positioned freely or moved dynamically during estimation. Notably, the same paradigm of using camera-frame point cloud observations has also demonstrated strong performance and generalization in policy learning \cite{ze2024generalizable,zhang2025canonical}.

For point clouds, extrinsic contact is annotated by having humans select contact points directly in 3D space. We will provide more details in Section~\ref{sec:data_collection_exp} on how this annotation process can be performed both quickly and reliably. Regardless of the contact type, this point-based annotation offers a consistent way to represent contacts, as shown in Fig.~\ref{fig:contact_representation}. We denote the set of $N$ annotated points as 
$ \{p^i_c\}_{i=1}^N$, where $p_c^i \in \mathbb{R}^3$ represent the $i$-th annotated point. 
Importantly, $N$ is not fixed; allowing it to vary across point clouds makes the annotation process highly flexible and simple. 

However, human annotations often exhibit non-uniform and sparse density across different regions, making them unsuitable as a dense supervision signal. To generate a uniform affordance representation from these annotated points, we adopt the following procedure, as shown in Fig.~\ref{fig:contact_representation}. For each point $p^k \in \mathbb{R}^3$, $k = 1, \dots, M$, in the point cloud captured by an RGB-D camera, we compute its minimum distance to the annotated set $\{p^i_c\}_{i=1}^N$:
$$
    d_k = \min\limits_{i=1,\dots,N} \|p^k - p_c^i\|_2.
$$
We then transform this distance into an affordance score using a Gaussian kernel: 
$$
    y_k = \exp\!\left(-\frac{d_k^2}{2\sigma^2}\right).
$$
Here, $\sigma$ controls the kernel bandwidth. Since the kernel value $y_k$ lies in $(0, 1]$, we further normalize it to match the affordance label range used during training. Specifically, we scale by a factor $s$, clamp to $[0, 1]$, and linearly rescale to the range $[-1,1]$:
$$
    a_k = 2 \cdot \mathrm{clamp}(s \cdot y_k, 0, 1) - 1,
$$
where the scaling factor $s$ controls the effective range before clamping.
This process yields an affordance value $a_k$ for each point $p^k$, where points in contact regions receive higher affordance values and those in non-contact regions receive lower values. As shown in Fig.~\ref{fig:contact_representation}, the affordance unifies contact and non-contact regions with minimal annotation effort and without requiring uniform human labeling.

\subsection{Multimodal Encoding}\label{sec:encoders}

We motivate our modality choices and describe how each input is encoded. In particular, we integrate four input streams: tactile sensing, end-effector rotation, force–torque sensing, and camera-frame point clouds.

\subsubsection{Selection of Modalities}
For tactile signals, we use marker displacement maps to capture the spatial distribution of shear across the finger, as shown in Fig.~\ref{fig:pipeline}. In our setup, two marker displacement maps are obtained from the parallel gripper’s two tactile sensors, which are GelSight Minis \cite{yuan2017gelsight}. Compared with other tactile representations such as raw images, depth maps, or binarized tactile maps, this representation provides a more unified and information-rich basis for contact estimation \cite{ma2021extrinsic}. The tactile data are structured in a tensor of size $(H, W, 2)$, where $H$ and $W$ denote the numbers of markers along the two axes, and the last dimension stores the $x$ and $y$ displacement components.

For the end-effector state, we use rotation rather than position or the full 6D pose. This design choice is motivated by the fact that rotation defines the coordinate frames of the force–torque and tactile signals, whereas position mainly introduces fixed geometric priors that are redundant in our setup. We represent rotation using quaternions. The force–torque modality is a 6D wrench from a wrist-mounted sensor. Point clouds are obtained in the camera frame from an RGB-D camera with variable placement, meaning the camera is not constrained to a single configuration during data collection and model deployment.

\subsubsection{Encoders}

Each modality is processed by a dedicated encoder to the latent space as illustrated in Fig.~\ref{fig:pipeline}.

\textbf{Tactile encoder:}  
Tactile inputs are processed by a transformer encoder tailored to marker displacements. Specifically, we patchify the $(H, W, 2)$ map via a convolutional projection with a stride equal to the patch size, flatten the resulting patches into a sequence of tokens, and apply a transformer encoder over this sequence \cite{dosovitskiy2020vit}. This yields a set of tactile tokens. In our setup, marker displacement maps from the two fingertips are independently encoded and then aggregated. This design suits marker displacements well: local correlations capture fine-grained contact patterns, while self-attention models long-range effects such as shear fields.

\textbf{End-effector rotation and force–torque encoders:}  
End-effector rotation and force–torque signals are each passed through lightweight MLPs to produce compact token sets.  

\textbf{Point cloud encoder:}  
Point clouds are processed with a PointNet encoder \cite{qi2017pointnet} to extract geometric features. These features are then pooled to a fixed number of tokens by averaging over downsampled subsets.  

\subsubsection{Shared Tokenization}
All modality-specific tokens are projected to a common embedding dimension and balanced to the same per-modality token count. For tactile, we flatten all patch tokens and apply a linear projection to a fixed number of tokens. For end-effector rotation and force-torque, linear projections reshape MLP features into the same token layout. For point clouds, the pooled encoder outputs are projected to the shared token dimension. This balancing ensures comparable contribution from each modality regardless of native resolution or feature size.

\subsection{Masked Multimodal Fusion} \label{sec:fusion}

We propose a masked multimodal fusion strategy to enhance the robustness and flexibility of multimodal representation. The key idea is to randomly mask a subset of latent tokens during training and replace them with a learnable mask token, which is then optimized jointly with the rest of the network, as shown in Fig.~\ref{fig:pipeline}. Both modality-specific tokens and mask tokens are then fed into the modality-agnostic transformer trunk.

This design brings three main benefits: \textbf{(i)} it simulates sensor dropouts or partial observations during training, forcing the model to leverage complementary signals; \textbf{(ii)} the learnable mask token provides a modality-agnostic prior that allows the transformer trunk to treat missing information properly rather than noisy or zero-padded inputs; and \textbf{(iii)} it ensures that the model naturally adapts to incomplete multimodal observations at deployment.

As described in Section~\ref{sec:encoders}, all modalities are encoded into a balanced tokenized representation. During training, we randomly mask a portion of these tokens according to a specified ratio, and in our experiments we set the masking ratio to 0.5. Masked positions are replaced with the learnable mask token, yielding a consistent placeholder that the modality-agnostic transformer trunk can interpret reliably across training iterations. The learnable mask token is updated through training.  The fused token sequence is then processed by the trunk, which performs cross-modal reasoning and produces a compact latent representation.

At deployment, random masking is disabled: all available modality tokens are passed directly into the transformer trunk. Thanks to this masking design, we can even discard one or more modalities at deployment while the system still functions reliably. If a modality is missing, its tokens are replaced with the learnable mask token, ensuring distributional consistency between training and inference. Consequently, the model remains reliable under sensor failures or missing inputs, without requiring retraining.

This design makes the overall system highly practical for real-world robotics. Since the trunk has been trained to process both modality tokens and mask tokens, it maintains stability under unpredictable sensing missing and supports flexible deployment across diverse hardware configurations.

\subsection{Efficient Sampling}\label{sec:efficient_sampling}

\begin{figure}[t]
\centering
\begin{overpic}[trim=0 0 0 0,clip, width=0.45\textwidth]{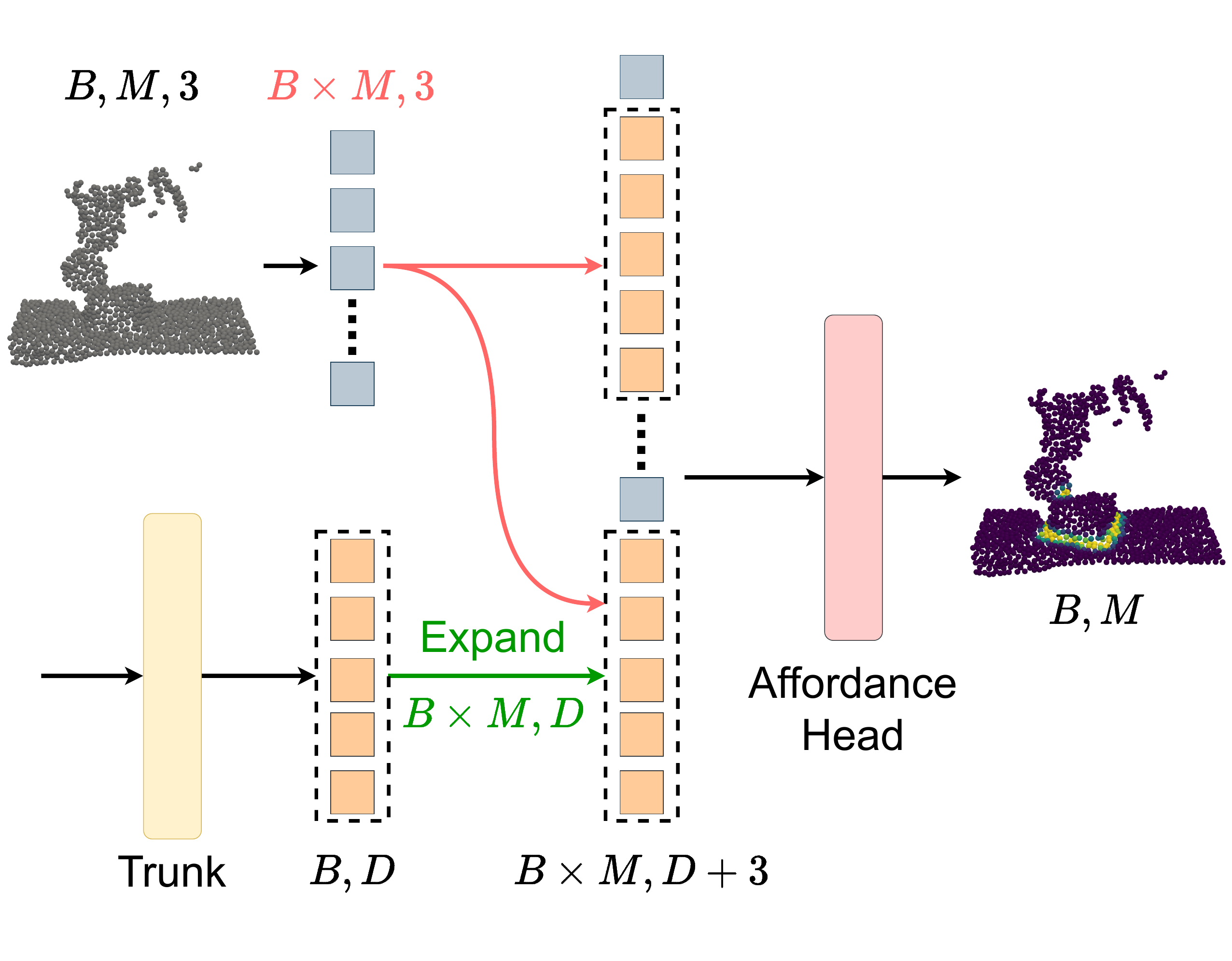}
\end{overpic}
\caption{Illustration of the sampling process. The gray blocks represent flattened sample points obtained from the camera-captured point cloud, while the orange blocks represent the multimodal feature generated by the trunk. Here, $B$ denotes the batch size, $M$ denotes the number of points in point cloud/sample points, and $D$ denotes the dimensionality of the multimodal feature. See Section~\ref{sec:efficient_sampling} for details.}
\label{fig:sampling}
\end{figure}

We present an efficient sampling strategy that decouples global multimodal fusion from point-wise affordance generation, thereby accelerating the forward pass. The transformer trunk shown in Fig.~\ref{fig:pipeline} and Fig.~\ref{fig:sampling} first produces a global multimodal feature of shape $(B, D)$, where $B$ is the batch size and $D$ the feature dimension. Meanwhile, the point cloud of size $M$ is retained as the set of sample points. The fused feature is then broadcast to these sample points and concatenated with their 3D coordinates, enabling lightweight point-wise inference. As illustrated in Fig.~\ref{fig:sampling}, this design pushes only lightweight computation to the point-wise head, resulting in a streamlined data flow: global fused feature $(B, D)$, broadcast to $(B \times M, D)$, concatenated with sample points $(B \times M, D{+}3)$, processed by the affordance head to $(B \times M, 1)$, and finally reshaped into the affordance map $(B, M)$.

\begin{figure}[ttt]
\centering
\begin{overpic}[trim=0 0 0 0,clip, width=0.38\textwidth]{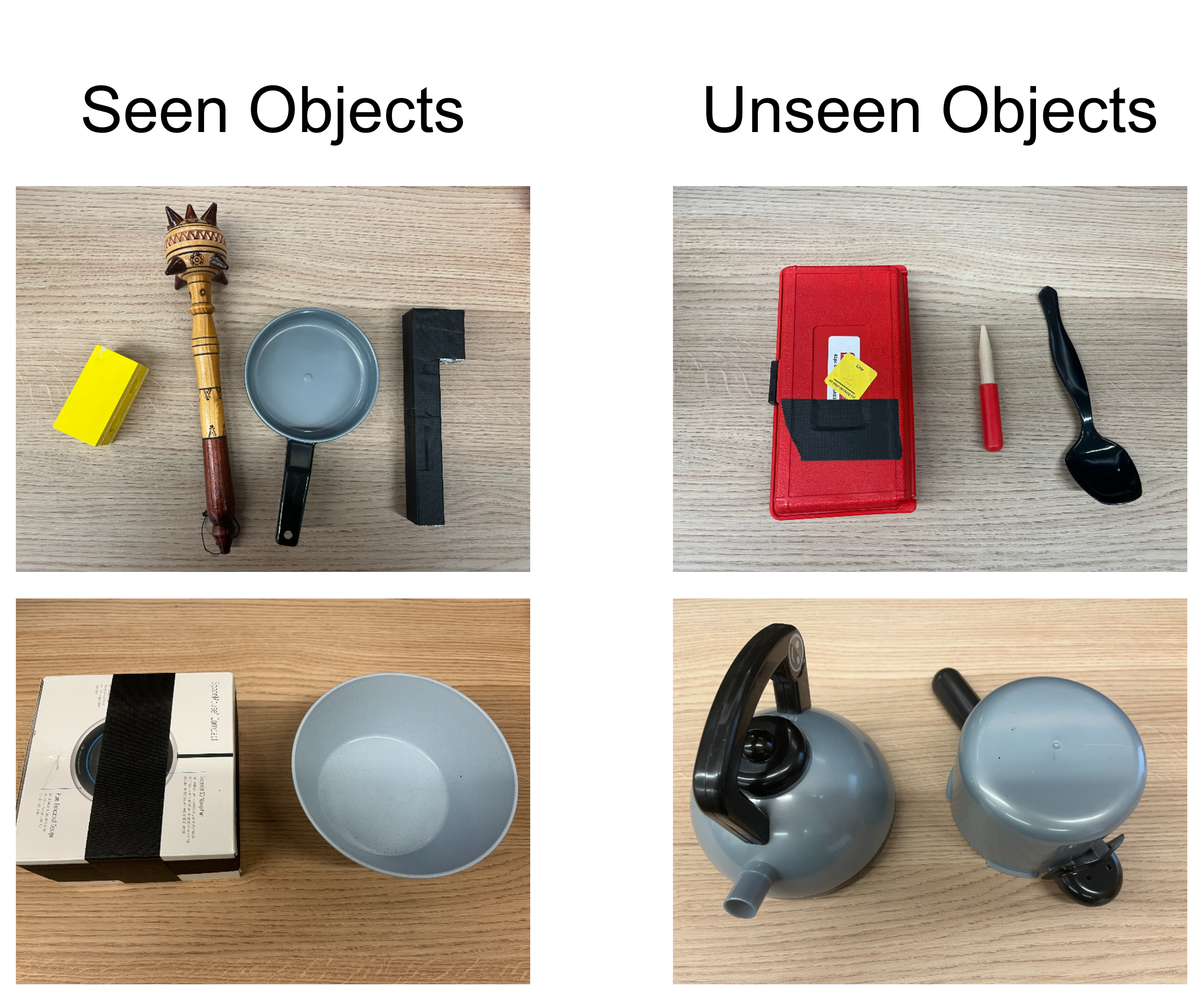}
\end{overpic}
\caption{Objects used in our experiments. The left set shows seen objects, while the right set shows unseen objects used only during validation.}
\label{fig:object}
\end{figure}

\section{Baselines}

In this section, we introduce three baselines which are benchmarked with \unic{} in experiments.

\subsection{\unic{} without Masked Fusion}
This baseline uses the same modalities, encoders, and sampling strategy as \unic{} model, but removes the masked multimodal fusion. During both training and inference, features from all modalities are directly concatenated and fed into the transformer trunk without any masking.

\subsection{End-to-end Regression}
End-to-end regression employs the same modalities and encoders but omits masked fusion, directly mapping the concatenated features to $L$ points as the predicted contact patch. Training is supervised with the Chamfer distance loss between predicted and ground-truth patches. To generate ground-truth patches, we first construct a dense affordance map, select points with positive affordance, and downsample them to $L$ points as ground-truth labels. For non-contact cases, all predicted points are set to zero~\cite{fu2025unitac}.

\subsection{Visual Estimation}
This baseline keeps the same sampling strategy but removes masked fusion and relies only on point cloud and end-effector rotation, excluding force–torque and tactile inputs.

\section{Experiments}

\begin{table*}[ttt]
\centering
\caption{Number of Episodes/Frames in Each Set. We split the dataset into training and validation sets with a ratio of 0.8:0.2.  See Section~\ref{sec:data_collection_exp} for more details about the dataset.}
\label{tab:dataset}
\begin{tabular}{|c|c|cc|ccc|}
\hline
       & All        & Train      & Valid    & All Contact & Single Contact & No Contact \\ \hline
Seen   & 1500/32274 & 1200/25751 & 300/6523 & 272/4476    & 131/1810       & 28/2047    \\
Unseen & 150/4555   & 0/0        & 150/4555 & 120/3250    & 60/1574        & 30/1305    \\ \hline
\end{tabular}
\end{table*}

\begin{figure*}[ttt]
\centering
\begin{overpic}[trim=0 0 0 0,clip, width=1\textwidth]{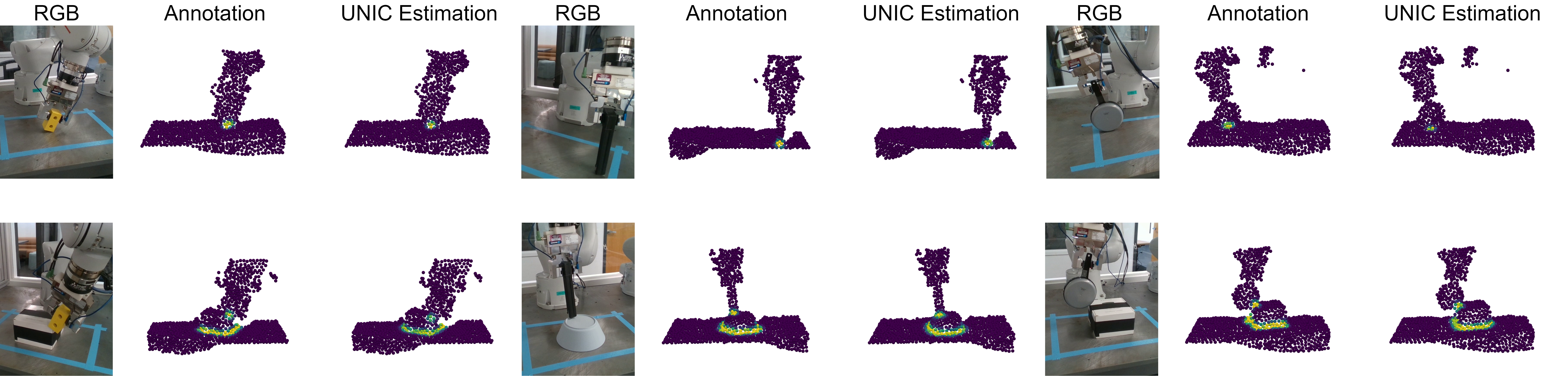}
\end{overpic}
\caption{\unic{} inference results on seen objects under unseen contact locations. \unic{} shows accurate extrinsic contact estimation across diverse contact configurations.}
\label{fig:seen_unic}
\end{figure*}

\begin{figure*}[ttt]
\centering
\begin{overpic}[trim=0 0 0 0,clip, width=1\textwidth]{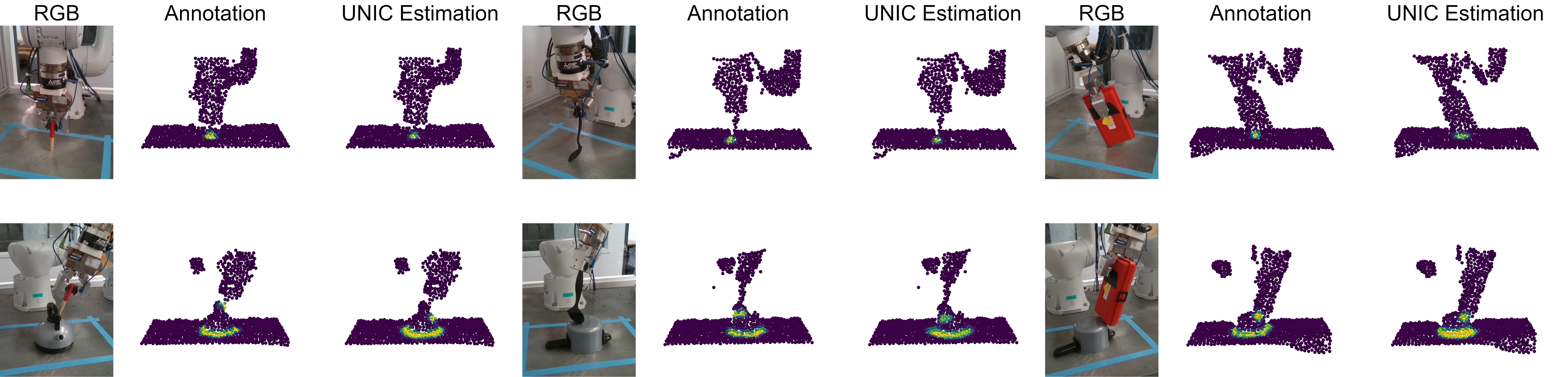}
\end{overpic}
\caption{\unic{} inference results on unseen objects, whose data were not used during training, as detailed in Sec.~\ref{sec:object_exp}. \unic{} demonstrates strong generalization to previously unseen objects.}
\label{fig:unseen_unic}
\end{figure*}

\begin{figure*}[ttt]
\centering
\begin{overpic}[trim=0 0 0 0,clip, width=1\textwidth]{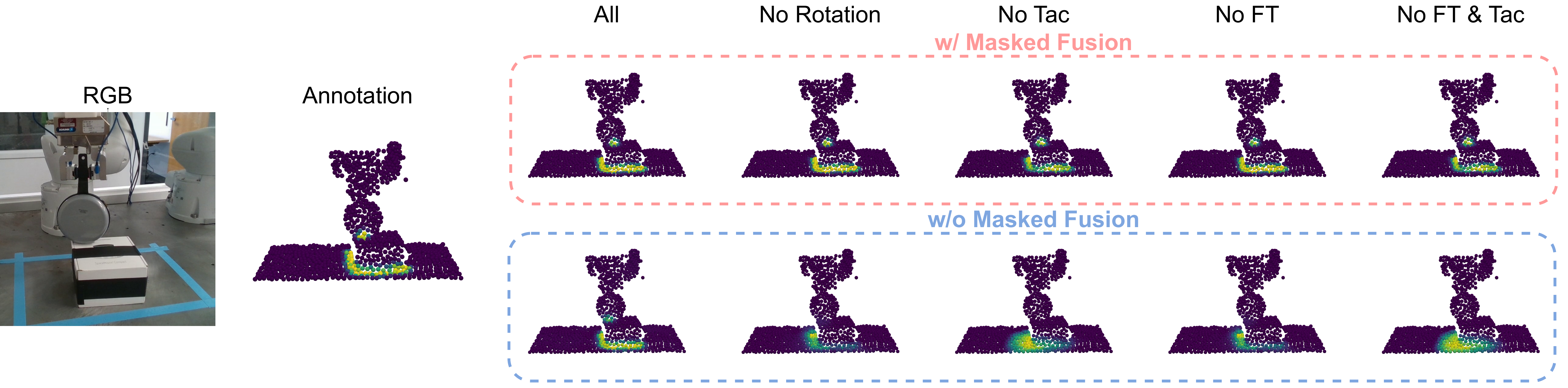}
\end{overpic}
\caption{Inference results with and without masked fusion under test-time modality removal. \unic{} with masked fusion demonstrates more robust predictions compared to the variant without masked fusion. This shows that masked fusion enables flexible test-time configurations of modality inputs without retraining the model, while maintaining performance.
}
\label{fig:blocking}
\end{figure*}

\begin{table*}[ttt]
\centering
\caption{Mean Absolute Error (Chamfer Distance, mm) of Contact Patch Estimation on the All Contact Validation Set.}
\label{tab:contact_patch}
\begin{tabular}{|c|ccccc|ccccc|}
\hline
\multirow{3}{*}{} & \multicolumn{5}{c|}{Seen Objects}                                                                                                                               & \multicolumn{5}{c|}{Unseen Objects}                                                \\ \cline{2-11} 
                  & \multicolumn{1}{c|}{\multirow{2}{*}{All}} & \multicolumn{4}{c|}{Test-time Modality Removal}                                                                                              & \multicolumn{1}{c|}{\multirow{2}{*}{All}} & \multicolumn{4}{c|}{Test-time Modality Removal}                 \\ \cline{3-6} \cline{8-11} 
                  & \multicolumn{1}{c|}{}                     & \multicolumn{1}{c|}{Rotation} & \multicolumn{1}{c|}{Tac} & \multicolumn{1}{c|}{FT} & FT \& Tac & \multicolumn{1}{c|}{}                      & Rotation   & Tac   & FT   & FT \& Tac \\ \hline
\unic{}              & \multicolumn{1}{c|}{9.6}                  & \textbf{9.4}            & \textbf{9.7}             & \textbf{9.7}             & \textbf{9.8}     & \multicolumn{1}{c|}{\textbf{33.4}}           & 32.5 & \textbf{33.5} & {33.5} & \textbf{32.9} \\
\unic{} w/o Masked Fusion  & \multicolumn{1}{c|}{\textbf{7.9}}         & 15.5                    & 199.2                    & 15.1                    & 210.2     & \multicolumn{1}{c|}{\textbf{33.4}}         & \textbf{28.4} & 215.4 & \textbf{27.9} & 230.3     \\
E2E Regression    & \multicolumn{1}{c|}{32.8}                 & 33.3                    & 36.9                     & 31.8                    & 34.9      & \multicolumn{1}{c|}{45.3}                  & 45.1 & 47.0 & 44.6 & 45.2 \\ 
Visual Estimation    & \multicolumn{1}{c|}{17.9}                  & 18.0                    & -                    & -                  & -      & \multicolumn{1}{c|}{{54.0}}       & 55.4 & - & - & -\\\hline
\end{tabular}
\end{table*}

\begin{table*}[ttt]
\centering
\caption{Mean Absolute Error (Distance, mm) of Contact Point Estimation on the Single Contact Validation Set. }
\label{tab:single_contact}
\begin{tabular}{|c|ccccc|ccccc|}
\hline
\multirow{3}{*}{} & \multicolumn{5}{c|}{Seen Objects}                                                                                                                               & \multicolumn{5}{c|}{Unseen Objects}                                                \\ \cline{2-11} 
                  & \multicolumn{1}{c|}{\multirow{2}{*}{All}} & \multicolumn{4}{c|}{Test-time Modality Removal}                                                                                              & \multicolumn{1}{c|}{\multirow{2}{*}{All}} & \multicolumn{4}{c|}{Test-time Modality Removal}                 \\ \cline{3-6} \cline{8-11} 
                  & \multicolumn{1}{c|}{}                      & \multicolumn{1}{c|}{Rotation} & \multicolumn{1}{c|}{Tac} & \multicolumn{1}{c|}{FT} & FT \& Tac & \multicolumn{1}{c|}{}                    & Rotation   & Tac   & FT   & FT \& Tac \\ \hline
\unic{}              & \multicolumn{1}{c|}{16.7}                  & \textbf{15.4}           & \textbf{16.2}            & \textbf{15.5}           & \textbf{16.8} & \multicolumn{1}{c|}{\textbf{23.4}}            & \textbf{23.0} & \textbf{25.3} & \textbf{23.8} & \textbf{25.1} \\
\unic{} w/o Masked Fusion  & \multicolumn{1}{c|}{\textbf{14.8}}         & 18.8                    & 118.4                    & 18.2                    & 128.0     & \multicolumn{1}{c|}{26.4}                & 27.8 & 125.7 & 28.6 & 132.4     \\
E2E Regression    & \multicolumn{1}{c|}{16.1}                  & 16.4                    & 23.1                     & 16.6                    & 23.0      & \multicolumn{1}{c|}{{25.3}}       & 24.6 & 26.7 & 24.7 & 27.2 \\
Visual Estimation    & \multicolumn{1}{c|}{22.2}                  & 22.2                    & -                    & -                  & -      & \multicolumn{1}{c|}{{40.0}}       & 41.0 & - & - & -\\\hline
\end{tabular}
\end{table*}

\begin{table*}[ttt]
\centering
\caption{Mean Absolute Error of Non-Contact Affordance Estimation. Since E2E regression does not predict affordance, it is omitted from this metric.}
\label{tab:non_contact}
\begin{tabular}{|c|ccccc|ccccc|}
\hline
\multirow{3}{*}{} & \multicolumn{5}{c|}{Seen Objects}                                                                                                                               & \multicolumn{5}{c|}{Unseen Objects}                                                   \\ \cline{2-11} 
                  & \multicolumn{1}{c|}{\multirow{2}{*}{All}} & \multicolumn{4}{c|}{Test-time Modality Removal}                                                                                              & \multicolumn{1}{c|}{\multirow{2}{*}{All}} & \multicolumn{4}{c|}{Test-time Modality Removal}                    \\ \cline{3-6} \cline{8-11} 
                  & \multicolumn{1}{c|}{}                     & \multicolumn{1}{c|}{Rotation} & \multicolumn{1}{c|}{Tac} & \multicolumn{1}{c|}{FT} & FT \& Tac & \multicolumn{1}{c|}{}                     & Rotation    & Tac   & FT    & FT \& Tac \\ \hline
\unic{}              & \multicolumn{1}{c|}{0.024}                & \textbf{0.050}          & \textbf{0.022}           & \textbf{0.088}          & \textbf{0.092}     & \multicolumn{1}{c|}{0.018}                & \textbf{0.012} & \textbf{0.014} & \textbf{0.058} & \textbf{0.076} \\
\unic{} w/o Masked Fusion  & \multicolumn{1}{c|}{\textbf{0.006}}       & 0.081                   & 0.092                    & 0.106                   & {0.104}     & \multicolumn{1}{c|}{\textbf{0.004}}  & 0.046 & 0.086 & 0.088 & 0.098 \\ 
Visual Estimation    & \multicolumn{1}{c|}{0.054}                  & 0.054                   & -                    & -                  & -      & \multicolumn{1}{c|}{{0.051}}       & 0.048 & - & - & -\\\hline
\end{tabular}
\end{table*}

\begin{table}[ttt]
\centering
\caption{Average inference time over $100$ forward passes under different test-time modality drop conditions.}
\begin{tabular}{lccccc}
\toprule
\textbf{All} & \textbf{No Rotation} & \textbf{No Tac} & \textbf{No FT} & \textbf{No FT \& Tac} \\
\midrule
0.0015~s & 0.0015~s & 0.0011~s & 0.0015~s & 0.0010~s \\
\bottomrule
\end{tabular}
\label{tab:inference_time}
\end{table}

In this section, we present our experimental setup and results. Our evaluation is designed to address the following key questions: 

\begin{enumerate}

\item Can \unic{} generalize to diverse contact locations, varying object configurations, different camera viewpoints, and even entirely unseen objects?

\item If models are trained with full modality inputs but some modalities are removed only at test time, can different methods still function effectively?

\item What insights can be drawn from the results regarding the relative importance of different modalities?

\item Is \unic{} capable of supporting real-time deployment?
\end{enumerate}

In our experiments, to ensure statistical significance of the results, we performed validation of all metrics every 10 epochs during training. For reporting, we averaged the results over the last 10 validations during training. This entire process was repeated with three different random seeds, and the results were obtained by averaging across these seeds.

\subsection{Setup}\label{sec:hardwaresetup}

\subsubsection{Hardware}
For data collection, we use a 6-DoF Mitsubishi MELFA robot equipped with a wrist-mounted force–torque sensor and a WSG-32 gripper, with each finger instrumented with a GelSight Mini. The robot operates under a stiffness controller to enable compliant interactions. An Intel RealSense D435 RGB-D camera is used to capture point cloud observations.

\subsubsection{Objects}\label{sec:object_exp}
As illustrated in Fig.~\ref{fig:object}, we employ two sets of objects for data collection. The seen objects are used to gather training data, and based on these objects we also construct a test set consisting of seen objects with unseen contact locations. The unseen objects are used exclusively to build another test set containing entirely novel objects that the model has never encountered. In Fig.~\ref{fig:object}, the top row shows the grasped objects, while the bottom row shows the objects placed on the table to provide contact surfaces.

\subsubsection{Data Collection}\label{sec:data_collection_exp}

All data streams were uniformly recorded at \SI{10}{\hertz}. Point clouds are captured with an flexible RGB-D camera, then cropped and downsampled to 1,024 points. The cropping is governed by a fixed set of parameters and applied identically across all data and inference runs. 

We use an episode as the basic unit, where each episode is a continuous data chunk lasting a few seconds. For each episode, we consider two cases: \textbf{(i)} contact at a fixed location, where the robot varies force and pose (e.g., rotation around the contact point). The contact point remains fixed within an episode but differs across episodes. \textbf{(ii)} no contact, where the end-effector moves freely. Although multimodal observations vary frame by frame, annotation is required only once per episode, as the fixed contact points generate affordance maps for all frames. Because contact locations vary across episodes and non-contact episodes involve unconstrained motion, episodes are naturally independent.

 To avoid data leakage, we split training and validation sets for seen objects at the episode level, with a ratio of 0.8:0.2. The resulting numbers of episodes and frames are reported in Table~\ref{tab:dataset}. For unseen objects, we collect both contact and non-contact episodes but use them only for validation. The whole validation set thus consists of: \textbf{(i)} unseen contact locations on seen objects, testing generalization to new configurations, and \textbf{(ii)} entirely unseen objects, testing generalization to new categories and shapes.

Each validation set (seen and unseen) is divided into all contact and no contact subsets, whose union forms the full set. Within the all contact subset, we further isolate single-object tabletop contacts for controlled evaluation, referred to as the single contact set. These partitions are constructed alongside annotation, ensuring consistency with semantic labels for analysis. Details of the dataset are provided in Table~\ref{tab:dataset}.

\subsection{Metrics}
We evaluate extrinsic contact estimation using three metrics, corresponding to the all contact, single contact, and no contact validation sets in Table~\ref{tab:dataset}.

In the all contact setting, contact is uniformly represented as a point cloud, allowing diverse forms of contact to be expressed as patches of points. The predicted patch is compared with the ground truth using Chamfer distance in mean absolute error. The results are summarized in Table~\ref{tab:contact_patch}.

The single contact set covers cases where the grasped object touches the table. A single representative point, obtained by averaging the contact patch, is compared with the ground-truth point using mean absolute error. The results are summarized in Table~\ref{tab:single_contact}.

The no contact set includes cases where the grasped object does not touch the environment. The model is expected to output an affordance map of all -1 values, and performance is measured by the mean absolute error between the predicted and ground-truth affordance maps. The results are summarized in Table~\ref{tab:non_contact}.

For both the all contact and single contact sets, contact patches are obtained by thresholding the affordance map: points with values greater than zero are selected as the contact point cloud.

\subsection{Results}

The quantitative results are summarized in Tables~\ref{tab:contact_patch}–\ref{tab:non_contact}. From these results, we draw the following conclusions:

\textbf{1. Performance on seen objects with unseen contact locations:} As shown in Tables~\ref{tab:contact_patch} and \ref{tab:single_contact}, \unic{} achieves a Chamfer distance error of 9.6 mm for contact patch estimation and a 16.7 mm distance error for single-object contact estimation. Importantly, these results are obtained under randomized camera viewpoints and without relying on any prior knowledge, demonstrating the impressive generalization capability of our approach. Representative visualization results are shown in Fig.~\ref{fig:seen_unic}. In our supplementary video, we present additional real-time inference results, including estimation on unseen contact locations, transitions from non-contact to contact states, and robust estimation under dynamic camera movements.

\textbf{2. Performance on unseen objects:} Even when deployed on entirely unseen objects, \unic{} exhibits acceptable performance, as shown in Tables~\ref{tab:contact_patch} and \ref{tab:single_contact}, with corresponding qualitative results illustrated in Fig.~\ref{fig:unseen_unic}. These results suggest that \unic{} can make reasonable predictions for previously unseen objects. This generalization stems from the fact that the input modalities used by \unic{}—point clouds and marker displacement maps—primarily capture geometric and interaction-related information, rather than relying on object-specific features. For objects with significantly different geometries, especially in multi-object contact scenarios, Fig.~\ref{fig:unseen_unic} shows that \unic{}’s predictions, while not always precise, remain spatially consistent and reasonable.

\textbf{3. Robustness of masked fusion on \unic{}:} As shown in Tables~\ref{tab:contact_patch} and \ref{tab:single_contact}, and Fig.~\ref{fig:blocking}, we also observe that \unic{} without masked fusion achieves higher performance than \unic{} when all modalities are present at test time. However, once a modality is removed during inference, its performance degrades substantially, with the drop being especially pronounced for high-dimensional inputs such as tactile marker displacement maps. This indicates that while masked fusion sacrifices some global performance, it greatly enhances the robustness of multimodal representations. Moreover, it introduces a new capability—allowing test-time modalities to be flexibly reconfigured without severely compromising performance—thereby making deployment both more robust and more adaptable.

\textbf{4. Role of tactile vs. force-torque:} From the evaluation on the all contact dataset, tactile marker tracking proves more beneficial for fine-grained estimation, as removing tactile input leads to a larger performance drop compared to removing force–torque, as shown in Table~\ref{tab:single_contact}. In contrast, force–torque signals play a more critical role in distinguishing whether contact occurs: on the no contact validation set, removing force–torque causes a significant degradation in performance, as shown in Table~\ref{tab:non_contact}. This is intuitive, since tactile marker tracking provides richer and more detailed feedback, implicitly encoding fine-grained interaction and spatial information that supports accurate estimation. Force–torque, on the other hand, is a low-dimensional signal that is more directly informative for detecting the presence or absence of contact.

\textbf{5. Comparison with baselines:} Both end-to-end regression and vision estimation perform worse than \unic{} and its ablations. For end-to-end regression, directly predicting contact patches without sampling makes it more difficult to capture the distribution of contacts, leading to substantially higher learning difficulty. For vision estimation, despite using the same architecture, the absence of force–torque and tactile inputs results in a pronounced drop in accuracy. This highlights the critical importance of multimodal inputs for precise extrinsic contact estimation.

\textbf{6. Inference time:} 
We measure the inference time of \unic{} on a single RTX 3080 GPU, both with all modalities and under different test-time modality drop conditions, as shown in Table~\ref{tab:inference_time}. Each reported value corresponds to generating a full contact affordance map from a point cloud of 1,024 points. \unic{} runs at over \SI{600}{\hertz}, demonstrating real-time efficiency.

\section{Discussion}

We introduced \unic{}, a unified multimodal framework for extrinsic contact estimation that eliminates reliance on prior knowledge and camera calibration. By directly leveraging multimodal inputs, \unic{} generalizes across diverse contact formations, remains robust under missing modalities, and demonstrates effectiveness across challenging scenarios.

Looking ahead, \unic{} can be advanced in three main directions. First, scalable multimodal data generation through simulation and sim-to-real transfer can overcome the limitations of human annotations and the scarcity of real-world data. Second, integrating vision \cite{simeoni2025dinov3} or vision-language foundation models \cite{yang2025qwen3} may strengthen multimodal alignment and contextual reasoning, enabling richer open-world understanding for contact estimation. Finally, coupling \unic{} with contact-rich manipulation policies \cite{higuera2023perceiving} could provide informative conditions for learning and control, paving the way for more adaptive and general-purpose robotic manipulation.

\bibliographystyle{IEEEtran}
\bibliography{reference}

@inproceedings{ma2021extrinsic,
  title={Extrinsic contact sensing with relative-motion tracking from distributed tactile measurements},
  author={Ma, Daolin and Dong, Siyuan and Rodriguez, Alberto},
  booktitle={2021 IEEE international conference on robotics and automation (ICRA)},
  pages={11262--11268},
  year={2021},
  organization={IEEE}
}

@inproceedings{higuera2023neural,
  title={Neural Contact Fields: Tracking Extrinsic Contact with Tactile Sensing},
  author={Higuera, Carolina and Dong, Siyuan and Boots, Byron and Mukadam, Mustafa},
  booktitle={2023 IEEE International Conference on Robotics and Automation (ICRA)},
  pages={12576--12582},
  year={2023},
  organization={IEEE}
}

@article{billard2019trends,
  title={Trends and challenges in robot manipulation},
  author={Billard, Aude and Kragic, Danica},
  journal={Science},
  volume={364},
  number={6446},
  pages={eaat8414},
  year={2019},
  publisher={American Association for the Advancement of Science}
}

@article{dosovitskiy2020vit,
  title={An Image is Worth 16x16 Words: Transformers for Image Recognition at Scale},
  author={Dosovitskiy, Alexey and Beyer, Lucas and Kolesnikov, Alexander and Weissenborn, Dirk and Zhai, Xiaohua and Unterthiner, Thomas and  Dehghani, Mostafa and Minderer, Matthias and Heigold, Georg and Gelly, Sylvain and Uszkoreit, Jakob and Houlsby, Neil},
  journal={ICLR},
  year={2021}
}

@inproceedings{qi2023general,
  title={General in-hand object rotation with vision and touch},
  author={Qi, Haozhi and Yi, Brent and Suresh, Sudharshan and Lambeta, Mike and Ma, Yi and Calandra, Roberto and Malik, Jitendra},
  booktitle={Conference on Robot Learning},
  pages={2549--2564},
  year={2023},
  organization={PMLR}
}

@inproceedings{hansen2022visuotactile,
  title={Visuotactile-rl: Learning multimodal manipulation policies with deep reinforcement learning},
  author={Hansen, Johanna and Hogan, Francois and Rivkin, Dmitriy and Meger, David and Jenkin, Michael and Dudek, Gregory},
  booktitle={2022 International Conference on Robotics and Automation (ICRA)},
  pages={8298--8304},
  year={2022},
  organization={IEEE}
}

@article{liu2025factr,
  title={FACTR: Force-Attending Curriculum Training for Contact-Rich Policy Learning}, 
  author={Jason Jingzhou Liu and Yulong Li and Kenneth Shaw and Tony Tao and Ruslan Salakhutdinov and Deepak Pathak},
  journal={arXiv preprint arXiv:2502.17432},
  year={2025}, 
}

@inproceedings{xue2025reactive,
  title     = {Reactive Diffusion Policy: Slow-Fast Visual-Tactile Policy Learning for Contact-Rich Manipulation},
  author    = {Xue, Han and Ren, Jieji and Chen, Wendi and Zhang, Gu and Fang, Yuan and Gu, Guoying and Xu, Huazhe and Lu, Cewu},
  booktitle = {Proceedings of Robotics: Science and Systems (RSS)},
  year      = {2025}
}

@inproceedings{bianchini2025vysics,
 title={Vysics: Object Reconstruction Under Occlusion by Fusing Vision and Contact-Rich Physics},
 author={Bibit Bianchini and Minghan Zhu and Mengti Sun and Bowen Jiang and Camillo J. Taylor and Michael Posa},
 year={2025},
 month={june}, 
 booktitle={Robotics: Science and Systems (RSS)},
 website={https://vysics-vision-and-physics.github.io/}, 
}

@article{lee2025vitascope,
  title={ViTaSCOPE: Visuo-tactile Implicit Representation for In-hand Pose and Extrinsic Contact Estimation},
  author={Lee, Jayjun and Fazeli, Nima},
  journal={arXiv preprint arXiv:2506.12239},
  year={2025}
}

@article{fu2025unitac,
  title={UniTac: Whole-Robot Touch Sensing Without Tactile Sensors},
  author={Fu, Wanjia and Li, Hongyu and He, Ivy X and Tellex, Stefanie and Sridhar, Srinath},
  journal={arXiv preprint arXiv:2507.07980},
  year={2025}
}

@article{yi2024visual,
  title={Visual-auditory Extrinsic Contact Estimation},
  author={Yi, Xili and Lee, Jayjun and Fazeli, Nima},
  journal={arXiv preprint arXiv:2409.14608},
  year={2024}
}

@article{kim2023simultaneous,
  title={Simultaneous tactile estimation and control of extrinsic contact},
  author={Kim, Sangwoon and Jha, Devesh K and Romeres, Diego and Patre, Parag and Rodriguez, Alberto},
  journal={arXiv preprint arXiv:2303.03385},
  year={2023}
}

@article{rodriguez2021unstable,
  title={The unstable queen: Uncertainty, mechanics, and tactile feedback},
  author={Rodriguez, Alberto},
  journal={Science Robotics},
  volume={6},
  number={54},
  pages={eabi4667},
  year={2021},
  publisher={American Association for the Advancement of Science}
}

@inproceedings{ota2024tactile,
  title={Tactile estimation of extrinsic contact patch for stable placement},
  author={Ota, Kei and Jha, Devesh K and Jatavallabhula, Krishna Murthy and Kanezaki, Asako and Tenenbaum, Joshua B},
  booktitle={2024 IEEE International Conference on Robotics and Automation (ICRA)},
  pages={13876--13882},
  year={2024},
  organization={IEEE}
}

@article{oller2024tactile,
  title={Tactile-driven non-prehensile object manipulation via extrinsic contact mode control},
  author={Oller, Miquel and Berenson, Dmitry and Fazeli, Nima},
  journal={arXiv preprint arXiv:2405.18214},
  volume={3},
  year={2024}
}

@article{aoyama2025few,
  title={Few-shot transfer of tool-use skills using human demonstrations with proximity and tactile sensing},
  author={Aoyama, Marina Y and Vijayakumar, Sethu and Narita, Tetsuya},
  journal={IEEE Robotics and Automation Letters},
  year={2025},
  publisher={IEEE}
}

@article{kim2024texterity,
  title={TEXterity--Tactile Extrinsic deXterity: Simultaneous Tactile Estimation and Control for Extrinsic Dexterity},
  author={Kim, Sangwoon and Bronars, Antonia and Patre, Parag and Rodriguez, Alberto},
  journal={arXiv preprint arXiv:2403.00049},
  year={2024}
}

@article{higuera2023perceiving,
  title={Perceiving extrinsic contacts from touch improves learning insertion policies},
  author={Higuera, Carolina and Ortiz, Joseph and Qi, Haozhi and Pineda, Luis and Boots, Byron and Mukadam, Mustafa},
  journal={arXiv preprint arXiv:2309.16652},
  year={2023}
}

@inproceedings{shirai2023tactile,
  title={Tactile tool manipulation},
  author={Shirai, Yuki and Jha, Devesh K and Raghunathan, Arvind U and Hong, Dennis},
  booktitle={2023 IEEE International Conference on Robotics and Automation (ICRA)},
  pages={12597--12603},
  year={2023},
  organization={IEEE}
}

@inproceedings{taylor2023object,
  title={Object manipulation through contact configuration regulation: multiple and intermittent contacts},
  author={Taylor, Orion and Doshi, Neel and Rodriguez, Alberto},
  booktitle={2023 IEEE/RSJ International Conference on Intelligent Robots and Systems (IROS)},
  pages={8735--8743},
  year={2023},
  organization={IEEE}
}

@article{yuan2017gelsight,
  title={Gelsight: High-resolution robot tactile sensors for estimating geometry and force},
  author={Yuan, Wenzhen and Dong, Siyuan and Adelson, Edward H},
  journal={Sensors},
  volume={17},
  number={12},
  pages={2762},
  year={2017},
  publisher={MDPI}
}

@inproceedings{qi2017pointnet,
  title={Pointnet: Deep learning on point sets for 3d classification and segmentation},
  author={Qi, Charles R and Su, Hao and Mo, Kaichun and Guibas, Leonidas J},
  booktitle={Proceedings of the IEEE conference on computer vision and pattern recognition},
  pages={652--660},
  year={2017}
}

@article{simeoni2025dinov3,
  title={Dinov3},
  author={Sim{\'e}oni, Oriane and Vo, Huy V and Seitzer, Maximilian and Baldassarre, Federico and Oquab, Maxime and Jose, Cijo and Khalidov, Vasil and Szafraniec, Marc and Yi, Seungeun and Ramamonjisoa, Micha{\"e}l and others},
  journal={arXiv preprint arXiv:2508.10104},
  year={2025}
}

@article{yang2025qwen3,
  title={Qwen3 technical report},
  author={Yang, An and Li, Anfeng and Yang, Baosong and Zhang, Beichen and Hui, Binyuan and Zheng, Bo and Yu, Bowen and Gao, Chang and Huang, Chengen and Lv, Chenxu and others},
  journal={arXiv preprint arXiv:2505.09388},
  year={2025}
}

@misc{luu2025manifeel,
  title={ManiFeel: Benchmarking and Understanding Visuotactile Manipulation Policy Learning},
  author={Quan Khanh Luu and Pokuang Zhou and Zhengtong Xu and Zhiyuan Zhang and Qiang Qiu and Yu She},
  year={2025},
  eprint={2505.18472},
  archivePrefix={arXiv},
  primaryClass={cs.RO},
  url={https://arxiv.org/abs/2505.18472},
}

@misc{xu2025unit,
  title={{UniT}: Data Efficient Tactile Representation with Generalization to Unseen Objects},
  author={Zhengtong Xu and Raghava Uppuluri and Xinwei Zhang and Cael Fitch and Philip Glen Crandall and Wan Shou and Dongyi Wang and Yu She},
  year={2025},
  eprint={2408.06481},
  archivePrefix={arXiv},
  primaryClass={cs.RO},
  url={https://arxiv.org/abs/2408.06481},
}

@article{ze2024generalizable,
  title={Generalizable humanoid manipulation with 3d diffusion policies},
  author={Ze, Yanjie and Chen, Zixuan and Wang, Wenhao and Chen, Tianyi and He, Xialin and Yuan, Ying and Peng, Xue Bin and Wu, Jiajun},
  journal={arXiv preprint arXiv:2410.10803},
  year={2024}
}

@misc{zhang2025canonical,
  title={Canonical Policy: Learning Canonical 3D Representation for Equivariant Policy},
  author={Zhiyuan Zhang and Zhengtong Xu and Jai Nanda Lakamsani and Yu She},
  year={2025},
  eprint={2505.18474},
  archivePrefix={arXiv},
  primaryClass={cs.RO},
  url={https://arxiv.org/abs/2505.18474},
}

\end{document}